\documentclass{article}
\usepackage{graphicx, float} 
\graphicspath{{images/}}
\usepackage[utf8]{inputenc}
\usepackage{amsmath, amssymb, amsthm}
\usepackage{booktabs}
\usepackage{array}
\usepackage{authblk}
\usepackage{hyperref}
\usepackage{ragged2e}
\usepackage{indentfirst}
\usepackage{caption}
\usepackage[top=1in, bottom=1in, left=1.5in, right=1.5in]{geometry}

\title{\textbf{An Enhancement of Jiang, Z., et al.’s Compression-Based Classification Algorithm Applied to News Article Categorization}}

\author[1]{Sean Lester C. Benavides}
\author[2]{Cid Antonio F. Masapol}
\author[3]{Jonathan C. Morano}
\author[4]{Dan Michael A. Cortez}
\affil[1,2,3,4]{Department of Computer Science, Pamantasan ng Lungsod ng Maynila, Manila, Philippines}
\affil[ ]{\textit{\small 
\textsuperscript{1}\href{mailto:slcbenavides2021@plm.edu.ph}{slcbenavides2021@plm.edu.ph},  
\textsuperscript{2}\href{mailto:cafmasapol2019@plm.edu.ph}{cafmasapol2019@plm.edu.ph},  
\textsuperscript{3}\href{mailto:jcmorano@plm.edu.ph}{jcmorano@plm.edu.ph},  
\textsuperscript{4}\href{mailto:dmacortez@plm.edu.ph}{dmacortez@plm.edu.ph}}}
\date{December 2024}

\begin{document}

\maketitle

\begin{abstract}
    This study enhances Jiang et al.'s compression-based classification algorithm by addressing its limitations in detecting semantic similarities between text documents. The proposed improvements focus on unigram extraction and optimized concatenation, eliminating reliance on entire document compression. By compressing extracted unigrams, the algorithm mitigates sliding window limitations inherent to gzip, improving compression efficiency and similarity detection. The optimized concatenation strategy replaces direct concatenation with the union of unigrams, reducing redundancy and enhancing the accuracy of Normalized Compression Distance (NCD) calculations. Experimental results across datasets of varying sizes and complexities demonstrate an average accuracy improvement of \textit{5.73\%}, with gains of up to \textit{11\%} on datasets containing longer documents. Notably, these improvements are more pronounced in datasets with high-label diversity and complex text structures. The methodology achieves these results while maintaining computational efficiency, making it suitable for resource-constrained environments. This study provides a robust, scalable solution for text classification, emphasizing lightweight preprocessing techniques to achieve efficient compression, which in turn enables more accurate classification
\end{abstract}

\justify

\section{Introduction}

In an era where the volume of digital information continues to grow exponentially, the ability to classify and organize text data efficiently has become a critical challenge. News organizations and content platforms face particular difficulties in categorizing vast quantities of articles accurately and quickly, which is essential for delivering relevant content to their users \cite{Zhu2022}. Existing text classification methods often require complex models, large datasets, and significant computational resources, making them expensive and inflexible \cite{Jianan2023}. These limitations emphasize the need for simpler, more efficient solutions that maintain accuracy while requiring less computing power.

This study focuses on enhancing the Compression-Based Classification Algorithm by Jiang, Z., et al. for detecting patterns and measuring similarities between text documents. The paper explored efficient and lightweight text classification techniques. They proposed using gzip compression and Normalized Compression Distance (NCD) to classify texts. Their research focused on improving compression-based text classification. They used data compression to measure document similarity \cite{Jiang2022}. While this method is efficient and does not require training, its fixed window size limits its ability to handle large or complex texts, impacting accuracy and efficiency. To improve document similarity measurement, this study developed a new concatenation strategy. Instead of directly combining words from two documents, it first identifies unique words from both and then combines them. This reduced redundancy and improved pattern recognition, leading to better classification results.

The original and improved algorithms were evaluated by comparing their accuracy and processing time using gzip compression with and without preprocessing to assess the improvement in similarity accuracy. The study used different datasets, including Scikit-Learn's "20 Newsgroups," Kaggle's "AG News" and "oh-r8-r52," and Hugging Face's "DBpedia," to comprehensively evaluate the enhanced algorithm's performance in text and document classification.

Using unigram extraction and compression improved accuracy by avoiding fixed window size limitations and focusing on key words, leading to more accurate similarity detection. This optimized approach allowed the algorithm to handle larger datasets and longer documents more effectively, making it more versatile for various text classification tasks, especially when resources are limited. This study showed how effective unigram extraction and compression are at enhancing lightweight text classification algorithms, especially for categorizing news articles.

\section{Literature Review}

\subsection{Jiang, Z., et al.’s Compression-Based Classification Algorithm}

Compression-based classification algorithms can identify similarities between two texts by examining patterns and repetitions within the data. These algorithms often employ techniques such as dictionary-based encoding, which replaces frequently occurring patterns with shorter representations. For example, if two texts share common phrases or sequences, the algorithm will efficiently encode these similarities, thereby using less space. This process naturally highlights the similarities between texts, as repeated patterns or sequences are compressed similarly. The ability to capture these similarities depends on the algorithm's design, with some algorithms being more proficient at recognizing and leveraging these patterns for effective compression \cite{Ozan2024}.

In the 2022 paper "Less is More: Parameter-Free Text Classification with Gzip" by Jiang, Z., et al., the authors introduced an innovative approach that serves as a straightforward, efficient, and versatile alternative to the commonly used deep neural networks (DNNs) for text classification tasks. This method integrates a lossless compression algorithm with a k-nearest-neighbor (k-NN) classifier, effectively bypassing the need for any pre-processing or training stages that are typically required in traditional classification methods. The proposed approach leverages the inherent ability of the Gzip compression algorithm to identify and exploit patterns within the text, which are then utilized by the k-NN classifier to perform the classification. This combination allows for a parameter-free classification process, simplifying the overall workflow and reducing computational overhead.

The k-nearest neighbors (kNN) algorithm has a computational complexity of  \(O(n^2)\), which means that its efficiency significantly decreases as the dataset size grows. This is due to the substantial computational resources needed to process larger datasets. As a result, the speed of kNN becomes a major bottleneck when working with very large datasets, limiting its practical usefulness and overall performance. This issue is especially pronounced in high-dimensional data or large-scale applications, where the computation time can become excessively long \cite{Jiang2023}.

\subsection{News Article Categorization}

News organizations and various online platforms encounter the complex task of efficiently organizing and classifying large volumes of news articles to ensure that users can easily access information that aligns with their interests. This classification process involves categorizing news articles into various fields such as technology, politics, sports, entertainment, business, and more \cite{Volety2024}. As the volume of news published online increases, it is crucial to categorize articles to help people easily locate the information they need. With news coming from both print and digital sources, effective methods for organizing and understanding it are essential. Storing unclassified news does not serve much purpose \cite{Singh2021}. To address this issue, a system that monitors news in specific areas and automatically organizes it into relevant categories is essential. News classification involves extracting key information from articles and categorizing them accordingly \cite{Dogra2022}.

Text classification involves categorizing text into predefined groups based on its content. This process assigns specific labels to text documents to identify their subject matter or intent \cite{Zhang2023}. It is often applied in commercial domains, including spam filtering, decision making, information extraction from raw data, and other uses. It is important for different aspects because it eliminates the need for manual data classification, which is both costly and time consuming \cite{Hassan2022}. Additionally, it is a core task that has greatly benefited from advancements in neural networks. However, these networks often require substantial computational resources, involving millions of parameters and large datasets with labeled data. This makes their use, optimization, and adaptation to out-of-distribution cases costly in practical applications \cite{Peters2023}.

\subsection{Gzip Compressor}

Originally developed in the mid-1990s, GZIP is based on the DEFLATE compression algorithm, which combines LZ77 and Huffman coding techniques to minimize file sizes. Although GZIP has been a standard for compression since its inception, its design is constrained by a 32 KB sliding window—a limitation stemming from the memory constraints of the time. As a result, GZIP is less efficient on modern hardware, which can easily access much larger amounts of memory \cite{Bullock2024}. \cite{Jiang2022} emphasized several limitations of the Gzip compression algorithm, one of which is its fixed sliding window size. This limitation restricts the algorithm's ability to effectively identify and match patterns that are distant from each other within large documents. As a result, Gzip's performance in compressing large files can be suboptimal when dealing with data that contains patterns or repeated sequences spread over long distances, as the fixed window size is unable to capture these distant patterns efficiently.

In text classification, Gzip identifies similarities by analyzing the extent of compression achievable when combining different texts, forming the basis for metrics like the Normalized Compression Distance (NCD). It is a versatile metric designed to identify all types of similarities between signals that other distance measures usually detect individually. The NCD values range from 0 to 1, where 0 signifies maximum similarity and 1 indicates no similarity \cite{Pascarella2022}.

\subsection{Unigram}

The essential preprocessing step for text is tokenization, which determines the level of granularity at which textual data is analyzed and generated. Tokenization involves breaking a text stream into smaller pieces called tokens. Traditionally, most NLP models have used words as their primary units of analysis. However, more recent methods have begun decomposing text into smaller units, such as character n-grams or even underlying bytes. UnigramLM, like WordPiece, uses language models to determine the best way to break words into smaller units. However, UnigramLM starts with a much larger vocabulary and then gradually reduces it using a probabilistic approach. This approach allows for multiple possible ways to break words, and it can even randomly choose different ways based on their probabilities. While not often used alone, UnigramLM is a key part of the SentencePiece toolkit \cite{Gasparetto2022}. 

A unigram is a specific type of n-gram, where n equals 1, and it refers to a sequence consisting of a single adjacent element extracted from a set of tokens or string elements. These elements can be letters, syllables, or complete words, depending on the application or the level of granularity desired. In simpler terms, a unigram represents an isolated unit, which could be a single character, syllable, or word, that is considered independently of its surrounding context. The concept of unigrams is widely utilized in various fields, including linguistics, speech recognition, and cryptography, where the frequency of unigrams plays a crucial role. In these domains, unigram frequencies help analyze and process large amounts of text by capturing the occurrence of individual words or symbols. This analysis calculates the probability of a word appearing based on previous words, aiding in language models, speech transcription, and cryptographic systems \cite{Jimoh2021}.

\section{Methods}

The Jiang, Z., et al.'s Compression-Based Classification Algorithm is enhanced with a novel approach of utilizing the concept of unigram extraction and compression during the preprocessing phase. By extracting unigrams, the algorithm can focus on compressing text containing only these features rather than the entire text document. This approach allows for more efficient processing and circumvents the limitations imposed by the sliding window size of compression algorithms.

\subsection{Unigram Extraction, Compression, and Concatenation}

A unigram refers to a sequence of a single adjacent element derived from tokens of string elements, which can be letters, syllables, or words. It is a type of n-gram where n = 1. Unigram frequencies are used in fields like linguistics, speech recognition, and cryptography to analyze text and calculate the probability of a word given the previous word \cite{Jimoh2021}. Traditional compression methods like gzip compress entire documents, which can be inefficient. Unigram extraction improves efficiency by compressing individual words instead of whole texts. This technique is used in preprocessing to enhance similarity measurement and classification.

\begin{figure}[H]
    \centering
    \includegraphics[width=4in]{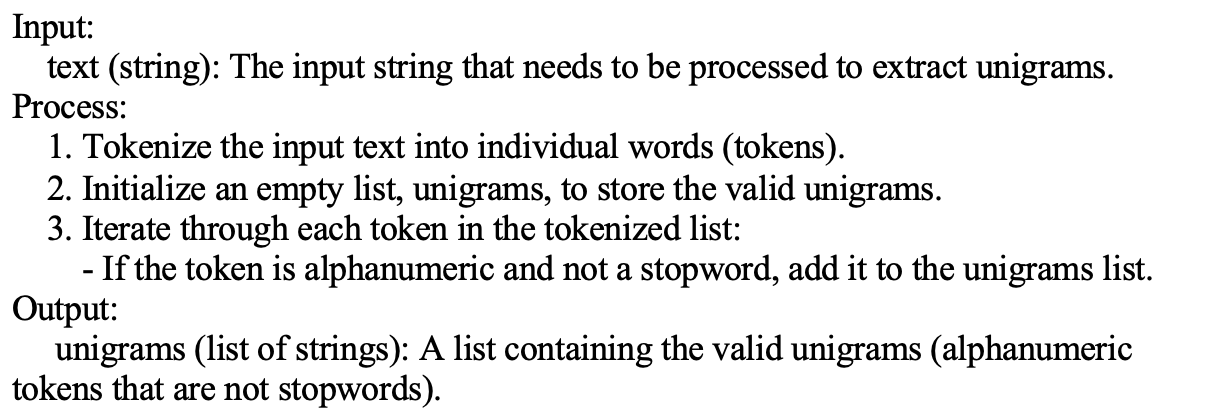}
    \caption{Unigram Extraction}
    \label{fig:unigram extraction}
\end{figure}

The unigram extraction operates by: 
\begin{enumerate}
    \item Tokenize the input text into individual words (tokens).
    \item Create an empty list called unigrams to store valid unigrams.
    \item For each token in the list of tokens:
    \begin{itemize}
        \item Check if the token is alphanumeric (letters or numbers).
        \item If condition is true, add the token to the unigrams list.
    \end{itemize}
\end{enumerate}

\begin{figure}[H]
    \centering
    \includegraphics[width=4in]{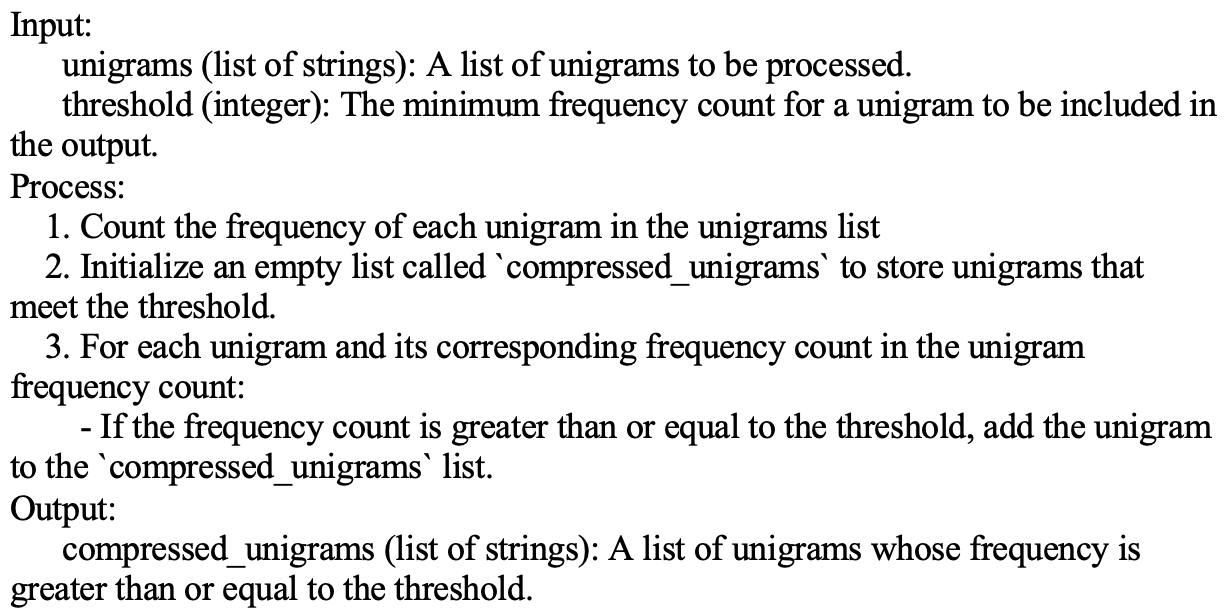}
    \caption{Unigram Compression}
    \label{fig:unigram compression}
\end{figure}

The unigram compression operates by: 
\begin{enumerate}
    \item Count the frequencies of each unigram in the unigrams list.
    \item Create an empty list called compressed\_unigrams to hold the unigrams that meet the threshold.
    \item For each unigram and count:
    \begin{itemize}
        \item Check if the count is greater than or equal to the threshold.
        \item If true, append the unigram to compressed\_unigrams.
    \end{itemize}
\end{enumerate}

Instead of concatenating documents, the union of their unigrams is compressed. This approach reduces text size, improving both compression efficiency. There are three possible scenarios:
\begin{itemize}
    \item Complete similarity: The union is half the size of the concatenation.
    \item Partial similarity: The union size depends on vocabulary overlap, often smaller than concatenation.
    \item Complete dissimilarity: The union equals the concatenation
\end{itemize}

\begin{figure}[H]
    \centering
    \includegraphics[width=4in]{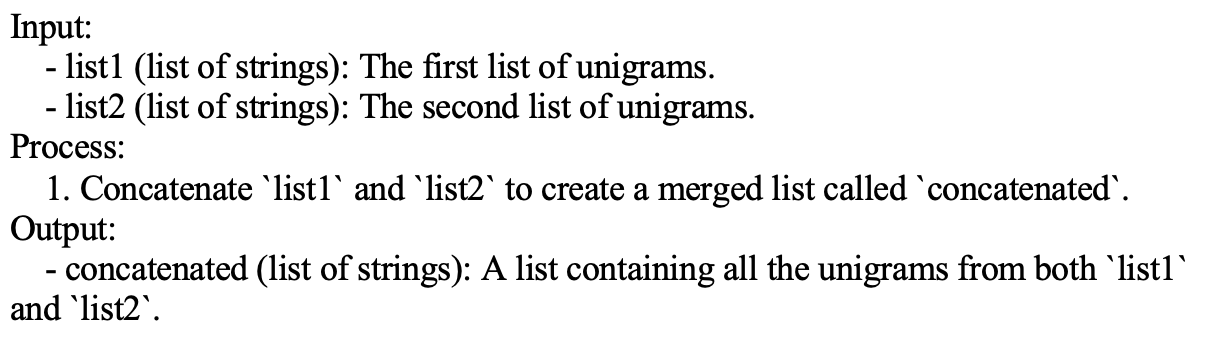}
    \caption{Unigram Concatenation}
    \label{fig:unigram concat}
\end{figure}

The unigram concatenation operates by: 
\begin{enumerate}
    \item Concatenate list1 and list2 to form a new list called concatenated.
    \item Return the concatenated list.
\end{enumerate}

\subsection{Enhanced  Jiang, Z., et al.'s Compression-Based Classification Algorithm}
\begin{figure}[H]
    \centering
    \includegraphics[width=4in]{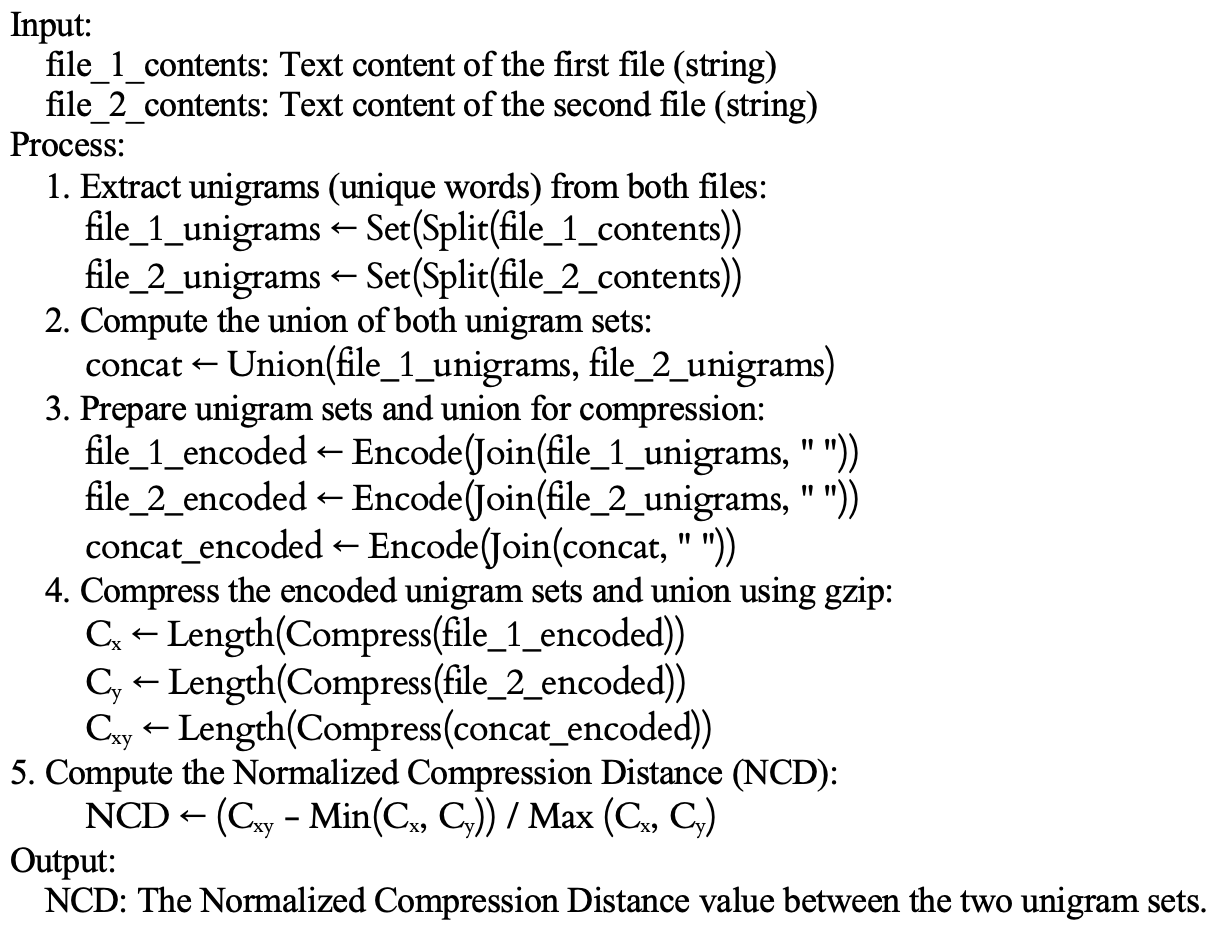}
    \caption{Enhanced  Jiang, Z., et al.'s Compression-Based Classification Algorithm}
    \label{fig:enhanced jiang}
\end{figure}
The enhanced Jiang, Z., et al.'s Compression-Based Classification Algorithm operates through six stages:
\begin{enumerate}
    \item Input of Documents
    \begin{itemize}
        \item Load or retrieve the text data of File 1
        \item Load or retrieve the text data of File 2
    \end{itemize}
    \item Extract Unique Words (Unigrams)
    \begin{itemize}
        \item Use the split() function to break each document into words
        \item Use the set() function to keep only unique words (unigrams) in each file
    \end{itemize}
    \item Compute the Union of Both Unigram Sets
    \begin{itemize}
        \item Combine the two unigram sets without duplicates using the union operation
    \end{itemize}
    \item Compress the Unigram Sets and Union
    \begin{itemize}
        \item Join the words in each set into a single string using spaces as separators
        \item Encode each string to bytes because the gzip.compress() function requires binary input
        \item Pass the encoded strings to the gzip.compress function \\
            Where: \\
                \(C_x = \) Length of the compressed version of file\_1\_encoded \\ 
                \(C_y = \) Length of the compressed version of file\_2\_encoded \\ 
                \(C_{xy} = \) Length of the compressed version of concat\_encoded
    \end{itemize}
    \item Compute the Normalized Compression Distance (NCD)
    \begin{itemize}
        \item Use the NCD formula to calculate the similarity between the two documents: 
           \[(NCD = \frac{C_{xy} - \min(C_x, C_y)}{\max(C_x, C_y)} \]
           
           Where: \\
            \(C_{xy}\) is the compressed size of the union. \\ 
            \(C_x\) is the compressed size of the first unigram set. \\
            \(C_y\) is the compressed size of the second unigram set.
        \item The NCD value will indicate the similarity between the two unigram sets \\
            Where: \\
            Low NCD: High similarity (unigrams overlap significantly) \\
            High NCD: Low similarity (unigrams differ significantly)
    \end{itemize}
\end{enumerate}

In the proposed Enhanced Jiang, Z., et al.'s Compression-Based Classification Algorithm, it enhances the process by extracting unique words, or unigrams, from each document during the preprocessing stage. By using sets to eliminate redundancy and computing the union of unigrams as the concatenated input, the algorithm focuses solely on word-level similarity. This preprocessing step, performed before compression, ensures that the NCD reflects the overlap of unique words while ignoring word frequency or text structure, resulting in a more concise and targeted similarity measure.

\subsection{News Article Categorization}

\begin{figure}[H]
    \centering
    \includegraphics[width=5in]{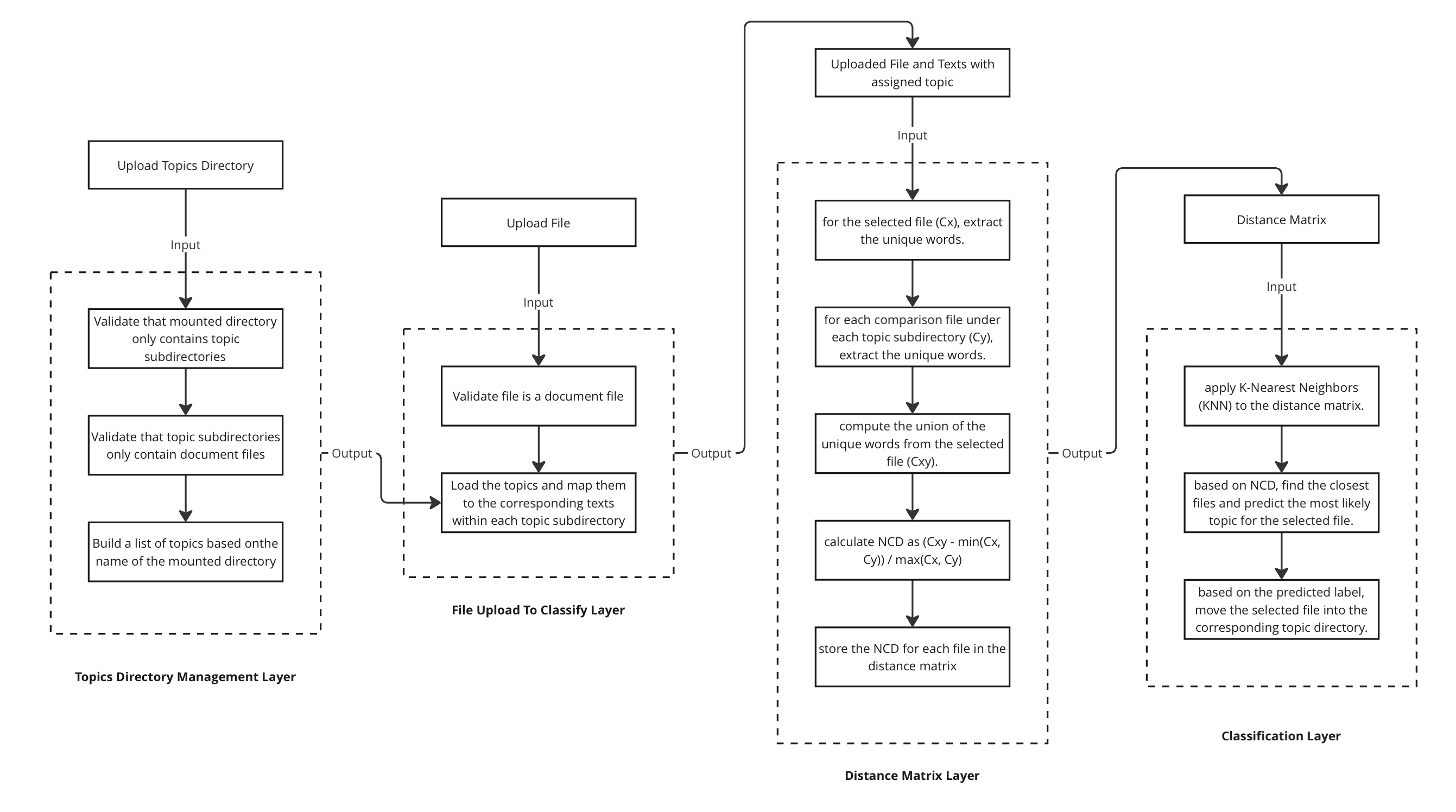}
    \caption{News Article Categorization System Architecture}
    \label{fig:systemarchi}
\end{figure}

Figure \ref{fig:systemarchi} illustrates a workflow for classifying uploaded documents into predefined topics, structured into four interconnected layers where each layer performs specific tasks to ensure accurate classification of the uploaded file.

\begin{enumerate}
    \item Topics Directory Management Layer: This layer uploads a directory of topic-specific subdirectories, validates the structure, and generates a list of topics for classification.
    \item File Upload to Classify Layer: This layer validates the uploaded document, loads the topic list, and maps the document to relevant topic subdirectories for classification.
    \item Distance matrix layer: This layer calculates the NCD between the uploaded file and files within each topic. It extracts unique words, computes their union, calculates NCD for similarity, and stores the results in a distance matrix.
    \item Classification Layer: The classification process uses the distance matrix and KNN algorithm to identify the closest files based on NCD values. It predicts the most likely topic for the uploaded file and moves it to the corresponding subdirectory.
\end{enumerate}
\subsection{Dataset Composition}

The enhanced algorithm was evaluated on diverse datasets, including newsgroups, news articles, academic papers, and classified documents, to assess its performance in high-volume, content-rich, and label-dense classification tasks.

\begin{enumerate}
    \item 20Newsgroup: A large-scale dataset featuring posts from 20 distinct newsgroups, providing longer texts and a higher label count, making it ideal for assessing gzip’s ability to manage content-rich documents.
    \item Ohsumed: A smaller dataset similar to 20Newsgroup for testing gzip’s efficiency on smaller datasets.
    \item AGNews: Consisting of short news articles categorized into four classes, this dataset tests gzip’s efficiency in handling high-volume, low-label tasks with concise documents.
    \item DBpedia (subset): Offering short documents with a moderate label count, this dataset helps evaluate gzip’s performance in handling classification tasks with a moderate level of complexity.
    \item R8 and R52: Subsets of the Reuters-21578 dataset, these datasets feature short to moderate text lengths with varied label counts, providing insights into gzip's accuracy across different label conditions in smaller datasets.
\end{enumerate}

By using a diverse range of text types, from news articles to academic papers and blogs, this study can ensure that the algorithm is thoroughly tested in a variety of classification contexts.

\section{Results}

This section presents the accuracy results of classification using NCD as the distance metric, with gzip as the compressor. It compares the baseline gzip approach to the improved method, which incorporates preprocessing.

\renewcommand{\arraystretch}{1} 
\setlength{\tabcolsep}{10pt} 
\begin{table}[H]
    \centering
    \captionsetup{justification=centering}
    \begin{tabular}{|c|c|c|c|}
    \hline
    \textbf{Dataset} & 
    \textbf{\begin{tabular}[c]{@{}c@{}}gzip on \\ Original  \\ Documents \\ Accuracy \\ (Jiang, Z., et al.’s)\end{tabular}} & 
    \textbf{\begin{tabular}[c]{@{}c@{}}gzip on \\ Preprocessed \\ Documents \\ Accuracy \\ (Enhanced)\end{tabular}} & 
    \textbf{\begin{tabular}[c]{@{}c@{}}Relative \\ Accuracy \\ Improvement \\(\%)\end{tabular}} \\ 
    \hline
    20Newsgroup & 0.52 & 0.57 & +9.6\% \\ 
    \hline
    AGNews & 0.88 & 0.91 & +3.4\% \\ 
    \hline
    DBpedia (subset) & 0.90 & 0.93 & +3.3\% \\ 
    \hline
    Ohsumed & 0.36 & 0.40 & +11.1\% \\ 
    \hline
    R8 & 0.90 & 0.93 & +3.3\% \\ 
    \hline
    R52 & 0.82 & 0.85 & +3.7\% \\
    \hline
    \end{tabular}
    \caption{Accuracy results comparing gzip compression on original documents vs. gzip compression on preprocessed documents}
    \label{table:table1}
\end{table}

The results in Table \ref{table:table1} demonstrate that preprocessing generally improves classification accuracy across all datasets, with varying degrees of improvement. The most significant improvements are observed in datasets with longer documents, such as Ohsumed and 20Newsgroup, where preprocessing helps mitigate the limitations of the sliding window in the gzip on original documents method. For example, Ohsumed experienced a \textbf{+11.1\%} relative improvement, suggesting that preprocessing reduced redundancy and enabled better compression of longer texts. Similarly, 20Newsgroup saw a \textbf{+9.6\%} relative improvement, indicating that preprocessing helped gzip handle lengthy documents more efficiently.

In contrast, datasets with relatively shorter documents, such as AGNews and DBpredia, showed more modest improvements. For AGNews, a \textbf{+3.4\%} relative increase was observed, reflecting that preprocessing had a smaller impact on simpler, shorter datasets. Likewise, DBpredia demonstrated a minimal accuracy improvement of \textbf{+3.3\%}, likely because the dataset consists of short documents where gzip performs reasonably well without preprocessing, as shown in the study of Jiang et al.

The results for R8 and R52 suggest that the number of labels in a dataset does not significantly influence the accuracy gains from preprocessing. Both datasets showed similar improvements, despite R52 having 52 categories compared to R8’s 8 categories. This finding suggests that document length and structure play a more critical role in determining the effectiveness of preprocessing than the number of categories.

\section{Discussion}

This study improves Jiang et al.'s compression-based classification algorithm by addressing its limitations in detecting semantic similarities. The key contributions are unigram extraction and optimized concatenation, which overcome the sliding window limitation of gzip. By focusing on compressing extracted unigrams instead of entire documents, and optimizing the concatenation of unigrams prior to compression, the study achieves significant improvements in classification accuracy, particularly for longer and more complex documents.

The results consistently show that preprocessing boosts classification accuracy across all datasets, especially in longer document datasets like Ohsumed and 20Newsgroup. These improvements are significant where gzip struggles due to its sliding window limitation. This aligns with Jiang et al.'s research, which found that longer documents face lower classification accuracy due to difficulty identifying patterns. This study extends that understanding by demonstrating that preprocessing can mitigate these issues, offering a scalable solution for handling longer, complex datasets. However, the degree of improvement remains dataset-dependent. Datasets with shorter documents, such as AGNews and DBpedia, displayed more modest improvements, supporting the notion that feature extraction methods are more beneficial for datasets with longer texts.

Interestingly, accuracy improvements in datasets with more categories, like R52, were similar to those in R8, despite the label differences. This suggests preprocessing benefits are more tied to document length and complexity than category count, as it reduces redundancy and enhances compression efficiency, particularly for longer documents. The proposed approach is ideal for resource-constrained environments, offering a lightweight alternative to intensive machine learning models. Its preprocessing improvements make it suitable for devices with limited computational power, such as mobile phones or IoT devices.

\section{Conclusion}

This study successfully demonstrated that unigram extraction and compression significantly enhance the performance of a compression-based classification algorithm for text categorization. By focusing on meaningful features through preprocessing, the algorithm's ability to detect semantic similarities was notably improved. The results showed that preprocessing steps, such as unigram extraction and optimized compression strategies, contributed to higher classification accuracy and efficiency across diverse datasets. The enhanced algorithm proves particularly effective for handling textual data in resource-constrained environments, offering a lightweight and computationally efficient solution. These advancements highlight the potential of compression-based methods in achieving robust and accurate text classification, particularly for applications like news article categorization.

\section{Limitations}

A limitation of this study is its reliance on controlled testing environments, which may not fully reflect the complexities of real-world news article categorization. Additionally, the study uses a relatively small set of six datasets, which may not represent the full diversity of real-world text classification tasks. Additionally, the study focuses solely on enhancing Jiang, Z., et al.'s Compression-Based Classification Algorithm for text classification in news article categorization. This narrow focus excludes the exploration of other potential applications, such as categorizing multimedia content like video or audio. Similarly, it does not extend to comparing the enhanced algorithm with alternative compressors or classification approaches, which could provide a broader understanding of its relative performance.

Additionally, the performance of the enhanced algorithm is potentially influenced by the specific hardware and computational resources available during testing. Since these conditions are not uniform across practical implementations, the algorithm's efficiency and accuracy might vary in environments with constrained resources or differing configurations. The study also does not account for the impact of evolving language patterns, unstructured data sources, or scalability challenges that could arise in real-world applications. Therefore, while the proposed enhancements demonstrate significant improvements under controlled conditions, further research is necessary to validate their robustness, scalability, and applicability across diverse datasets, evolving content types, and dynamic operational variables.

\bibliographystyle{IEEEtran}
\bibliography{references}
\end{document}